\documentclass[10pt,a4paper,twocolumn]{article}
\usepackage[margin=21mm,top=22mm,bottom=23mm]{geometry}
\usepackage[T1]{fontenc}
\usepackage{lmodern,microtype,amsmath,amssymb,booktabs,array,graphicx,xcolor,hyperref,longtable}
\hypersetup{colorlinks=true,urlcolor=blue!55!black,citecolor=blue!45!black,linkcolor=blue!45!black}
\usepackage{authblk}
\usepackage{enumitem}
\setlist[itemize]{nosep,leftmargin=*}
\title{\bfseries Transformer-Guided Short-Horizon Predictive Control for Multirotor Landing on Moving Offshore Decks: Computation and Terminal Feasibility}
\author{Athanasios Zeris\thanks{Independent Researcher, Athens, Greece. Email: \href{mailto:athzeris@gmail.com}{athzeris@gmail.com}. ORCID: \url{https://orcid.org/0009-0002-6907-2400}. Code: \url{https://github.com/AthanasiosZeris/UAV_Offshore_Landing_MPC}.}}
\date{}
\begin{document}
\flushbottom
\twocolumn[{
\begin{@twocolumnfalse}
\begin{center}
{\LARGE\bfseries Transformer-Guided Short-Horizon Predictive Control for Multirotor Landing on Moving Offshore Decks: Computation and Terminal Feasibility\par}
\vspace{2mm}
{\large Athanasios Zeris\textsuperscript{*}\par}
\vspace{1mm}

\end{center}
\vspace{2mm}
\noindent\textbf{Abstract.} 
Landing a multirotor drone on a moving offshore deck requires tracking a moving target while respecting vehicle dynamics and actuator limits. This study evaluates model predictive control (MPC) implemented by sequential convex programming (SCP), with Haar-basis control corrections and Transformer-generated initial guesses. Full-horizon and short-decision-horizon controllers are compared under common terminal position and relative-velocity criteria. The shorter formulation reduces computation time but decreases the number of cases satisfying the prescribed terminal position and velocity requirements; larger Transformer training sets do not remove that limitation. Fixed, height-dependent and closing-time-dependent attitude-alignment weights are also compared without claiming a general improvement in landing feasibility. In a matched closed-loop case, sensitivity condensation and forward-difference derivatives reduce the median computation time by 54.9\% while retaining close numerical agreement with the reference controller. Most updates nevertheless exceed the prescribed sampling interval. The simulations end at a clearance-offset target above the deck and do not model touchdown contact, execution-time delays or flight tests. The findings concern computational trade-offs and simulated terminal tracking, not demonstrated real-time autonomous landing.

\par\vspace{1mm}
\vspace{3mm}\noindent\textbf{Nomenclature}\par\vspace{1mm}
\footnotesize
\begin{tabular}{@{}p{.12\linewidth}p{.34\linewidth}p{.12\linewidth}p{.33\linewidth}@{}}
$\mathbf p,\mathbf v$ & UAV position and velocity & $q,\boldsymbol\omega$ & Quaternion and body angular velocity\\
$\mathbf x,\mathbf u$ & UAV state and thrust--torque input & $T,\boldsymbol\tau$ & Total thrust (N) and body torques (N\,m)\\
$m,J$ & Vehicle mass and inertia matrix & $\mathbf e_3,\ell$ & Vertical unit vector and deck-point offset\\
$R(q),R_d$ & UAV and deck rotations & $\mathbf p_d,\mathbf v_d$ & Deck target position and velocity\\
$\boldsymbol\Omega_d,\boldsymbol\Omega_d^b$ & Deck angular velocity in inertial and deck frames & $h_c,\mathbf n_d$ & Clearance and deck normal\\
$\mathbf p_r,\mathbf v_r$ & Clearance-offset reference position and velocity & $\mathbf p_f,\mathbf v_f$ & UAV terminal position and velocity\\
$\mathbf p_{r,f},\mathbf v_{r,f}$ & Terminal reference position and velocity & $e_f$ & Terminal tracking residual vector\\
$f,r$ (subscripts) & Terminal-time evaluation, reference & $\mathbf r,\dot{\mathbf r}$ & Deck-frame relative position and velocity\\
$T_s,T_F$ & Update period and terminal time & $H,P,B$ & Prediction steps, optimized steps, basis size\\
$\eta_j,H_j$ & Deck motion coordinate and response filter & $\Phi_{j:i}$ & State-transition product\\
$A_k,B_k$ & Local dynamics Jacobians & $S_k,D_k$ & Condensed state/control sensitivities\\
$\Psi,\delta c$ & Haar basis and coefficient correction & $\rho_x,\rho_u$ & SCP trust-region radii\\
$\mathbf e_R,\mathbf e_\omega$ & Deck-relative attitude/rate errors & $U_{\rm ref}$ & Transformer-proposed controls\\
FAA, PFAA & Attitude-weighting schemes & SCP, MPC & Sequential convex programming, predictive control\\
RTG, CTG & Return- and constraint-to-go & RMSE & Root mean square error
\end{tabular}
\normalsize

\vspace{4mm}
\end{@twocolumnfalse}
}]
\begingroup\renewcommand\thefootnote{*}\renewcommand\theHfootnote{authornote}\makeatletter\def\Hy@footnote@currentHref{authornote}\makeatother\footnotetext{Independent Researcher, Athens, Greece. Email: \href{mailto:athzeris@gmail.com}{athzeris@gmail.com}. ORCID: \url{https://orcid.org/0009-0002-6907-2400}. Code: \url{https://github.com/AthanasiosZeris/UAV_Offshore_Landing_MPC}.}\endgroup
\section{Introduction}
Model predictive control (MPC) repeatedly predicts the future motion of the controlled system, optimizes a constrained sequence of inputs and applies its first input before updating the plan. For landing on a moving offshore platform, the terminal target varies with deck heave, roll and pitch, and its velocity and angular rate matter alongside position. Recent work on agile landing on moving boats \cite{prochazka2024} and UAV operation over turbulent open water \cite{gupta2023} complements earlier studies of cooperative boat--UAV landing \cite{persson2019} and carrier-deck heave compensation \cite{koo2015}. In particular, Proch\'azka and coauthors employ phase-dependent penalties and deck-attitude synchronization, including real-world onboard validation. Consequently, attitude alignment itself is not claimed as a new concept here. Related UAV predictive-control and trajectory-optimization methods provide broader context \cite{ru2017,rohr2021,torrente2021,kelly2017}; they are not numerical baselines in the present experiments.

The broader maritime control literature has examined predictive ship-motion control and constraints \cite{sotnikova2012}, disturbance compensation for heading control \cite{lisun2012}, and robust roll/yaw control \cite{jin2020}. These studies motivate the treatment of moving and disturbed platforms but do not constitute validation of the UAV model used here. The present deck trajectory is prescribed or generated by a deterministic illustrative spectral realization, not a vessel-calibrated response-amplitude-operator model.

Finite-horizon predictive control and its computational trade-offs are established in standard treatments \cite{rawlings2017,borrelli2017}. Structure-exploiting online optimization can greatly reduce MPC computation \cite{wangboyd2010}, while move blocking reduces control degrees of freedom at the cost of possible feasibility or performance losses \cite{cagienard2007}. Different model granularity and nonuniform prediction grids offer another approach \cite{brudigam2021}. The Haar representation examined here is a reduced input-correction basis, not a claim to invent move blocking or nonuniform-horizon MPC. Wavelet-based trajectory representations have also been used for trajectory optimization in autonomous racing \cite{nam2026}.

Learning-assisted initialization is distinct from changing the optimization problem. Transformer-based trajectory proposals have been investigated for spacecraft rendezvous \cite{guffanti2024}; the sequence-model MPC architecture motivating this implementation is described by Celestini and coauthors \cite{celestini2024}. Transformer-accelerated iterative LQR provides a different computational route \cite{quattro2025}. An earlier study by the present author compared shifted and Transformer-based warm starts for MPC of a multirotor carrying a cable-suspended payload \cite{zeris2026payload}. The present study instead considers landing on a moving offshore deck, with emphasis on computation time and terminal tracking requirements. Two Transformer implementations must be distinguished: the original autoregressive trajectory model predicts physical actions, whereas the later HT10/FAA benchmark network predicts Haar coefficient arrays which are decoded into physical control proposals. In both cases SCP optimizes a subsequent correction. Neither representation guarantees terminal feasibility.

The experiments ask three questions: (1) When fewer future controls are optimized, how often does the planned approach still meet its final position and velocity requirements? (2) Does giving the Transformer more training examples improve those requirements, rather than only its action-prediction error? (3) Can simpler derivative calculations and a smaller optimization problem make the controller faster without noticeably changing the resulting simulated trajectory? We distinguish a single planning calculation from a closed-loop simulation that repeatedly replans. Computation time (``wall time'') means elapsed clock time for the timed controller calculation, including model evaluations, derivative construction and optimization, rather than simulated flight time. It is compared with the nominal $T_s=0.2$~s update interval.

The contributions are (i) a numerical comparison of Transformer initialization and Haar-parameterized SCP; (ii) terminal-aware fixed-tail short-decision-horizon control with FAA/PFAA weight variants; and (iii) a matched numerical evaluation of sensitivity-condensed SCP, forward-difference Jacobians and their combination. The results do not establish contact safety, general feasibility guarantees or real-time onboard execution.

For a moving landing target, position accuracy alone is insufficient: the vehicle must also approach with a small velocity relative to the target. A computationally faster controller is therefore useful only if its predicted and simulated terminal motion remains consistent with the specified tracking requirements.

\noindent\textbf{Code availability.} The GitHub repository for this study is \url{https://github.com/AthanasiosZeris/UAV_Offshore_Landing_MPC}. Public release of the Python source code, trained network checkpoints and reproducibility notebooks is planned through this repository.

\subsection{Landing-state criterion and scope of the comparisons}
The terminal residual is evaluated as four physically distinct components,
\begin{equation}
 e_f=\begin{bmatrix}(p_f-p_{r,f})^\top &(v_f-v_{r,f})^\top&e_{R,f}^\top&e_{\omega,f}^\top\end{bmatrix}^\top.
\end{equation}
The subscript $f$ denotes evaluation at the prescribed terminal time $T_F$, and the subscript $r$ identifies the clearance-offset reference. The numerical first-update feasibility label requires $\|p_f-p_{r,f}\|<0.50$~m and $\|v_f-v_{r,f}\|<0.30$~m/s; these are retrospective evaluation thresholds, not validated contact-safety limits. We also report how closely the vehicle matches the deck orientation and angular velocity. These two errors are not part of the binary feasibility label. Meeting the position and velocity thresholds means that the simulated approach reaches the specified target region; it does not mean that the vehicle has touched down safely. The terminal time is prescribed: this controller does not select a favorable sea-wave phase or search over landing windows. In the shrinking-horizon experiments the final physical time remains $T_F$; it is not reset to four seconds after each update.

\section{Multirotor and moving-deck model}
The control input consists of the total thrust $T$ (the sum of the rotor thrust forces) and the three body-axis torques $\boldsymbol\tau=[\tau_x,\tau_y,\tau_z]^\top$. The symbol $T$ is a force in newtons; $T_s$ and $T_F$ denote times.
The 13-component state and four-component control are
\begin{equation}
\mathbf x=[\mathbf p^\top,\mathbf v^\top,q^\top,\boldsymbol\omega^\top]^\top,\qquad
\mathbf u=[T,\tau_x,\tau_y,\tau_z]^\top .
\end{equation}
With scalar-first unit quaternion $q$ and body-to-inertial rotation $R(q)$, the modeled dynamics are
\begin{align}
\dot{\mathbf p}&=\mathbf v,&
\dot{\mathbf v}&=\frac{T}{m}R(q)\mathbf e_3-g\mathbf e_3,\\
\dot q&=\frac12q\otimes[0,\boldsymbol\omega],&
J\dot{\boldsymbol\omega}&=\boldsymbol\tau-\boldsymbol\omega\times J\boldsymbol\omega.
\end{align}
The model uses $m=2.0$~kg, $J=\mathrm{diag}(0.035,0.035,0.060)$~kg\,m$^2$, $g=9.81$~m/s$^2$, $0\leq T\leq35$~N, $|\tau_x|,|\tau_y|\leq0.08$~N\,m and $|\tau_z|\leq0.04$~N\,m. A fourth-order Runge--Kutta map $F_{T_s}$ propagates the state with quaternion renormalization. State perturbations for SCP use a 12-dimensional attitude-tangent representation rather than unconstrained perturbations to the four quaternion coordinates.

Let $\mathbf p_s,\mathbf v_s$ denote a ship reference point and velocity, $R_d$ the deck attitude, $\boldsymbol\ell$ the fixed landing-point offset and $\boldsymbol\Omega_d$ the deck angular velocity. The deck target and its normal are
\begin{equation}
\mathbf p_d=\mathbf p_s+R_d\boldsymbol\ell,\quad
\mathbf v_d=\mathbf v_s+\boldsymbol\Omega_d\times R_d\boldsymbol\ell,\quad
\mathbf n_d=R_d\mathbf e_3 .
\end{equation}
The \emph{pre-contact} reference is displaced $h_c=0.20$~m along the normal:
\begin{equation}
\mathbf p_r=\mathbf p_d+h_c\mathbf n_d,\qquad
\mathbf v_r=\mathbf v_d+h_c(\boldsymbol\Omega_d\times\mathbf n_d).
\end{equation}
The terminal tracking errors include $\mathbf p(T_F)-\mathbf p_r(T_F)$, $\mathbf v(T_F)-\mathbf v_r(T_F)$, relative attitude $\mathbf e_R$ and relative angular rate $\mathbf e_\omega$. In deck coordinates, $\mathbf r=R_d^\top(\mathbf p-\mathbf p_d)$ and
$\dot{\mathbf r}=R_d^\top(\mathbf v-\mathbf v_d)-(R_d^\top\boldsymbol\Omega_d)\times\mathbf r$.
The target is $\mathbf r=[0,0,h_c]^\top$, not impact with the physical deck.

\subsection{Rigid-body model assumptions and integration}
The modeled UAV is a rigid quadrotor with constant mass and diagonal inertia. The continuous dynamics are
\begin{align}
\dot p&=v,& \dot v&=\frac{T}{m}R(q)e_3-ge_3,\\
\dot q&=\frac12 q\otimes[0,\omega],&J\dot\omega&=\tau-\omega\times(J\omega).
\end{align}
For a scalar-first quaternion $q=(q_0,q_v)$, the body-to-inertial rotation is
\begin{equation}
 R(q)=(q_0^2-q_v^\top q_v)I+2q_vq_v^\top+2q_0[q_v]_\times.
\end{equation}
The nominal hover input is $u_h=[mg,0,0,0]^\top$. The simulation uses $m=2.0$~kg, $J=\mathrm{diag}(0.035,0.035,0.060)$~kg\,m$^2$, and $g=9.81$~m/s$^2$. Actuator magnitude limits are $0\leq T\leq35$~N, $|\tau_x|,|\tau_y|\leq0.08$~N\,m and $|\tau_z|\leq0.04$~N\,m. No separate experimentally validated motor allocation or slew-rate certification is implied by these command limits. Fourth-order Runge--Kutta integration advances the physical model, with unit-quaternion normalization. The finite-difference Jacobians are computed from this discrete map, rather than substituting a different plant for the local optimization.

\begin{figure*}[t]
\centering\includegraphics[width=.91\textwidth]{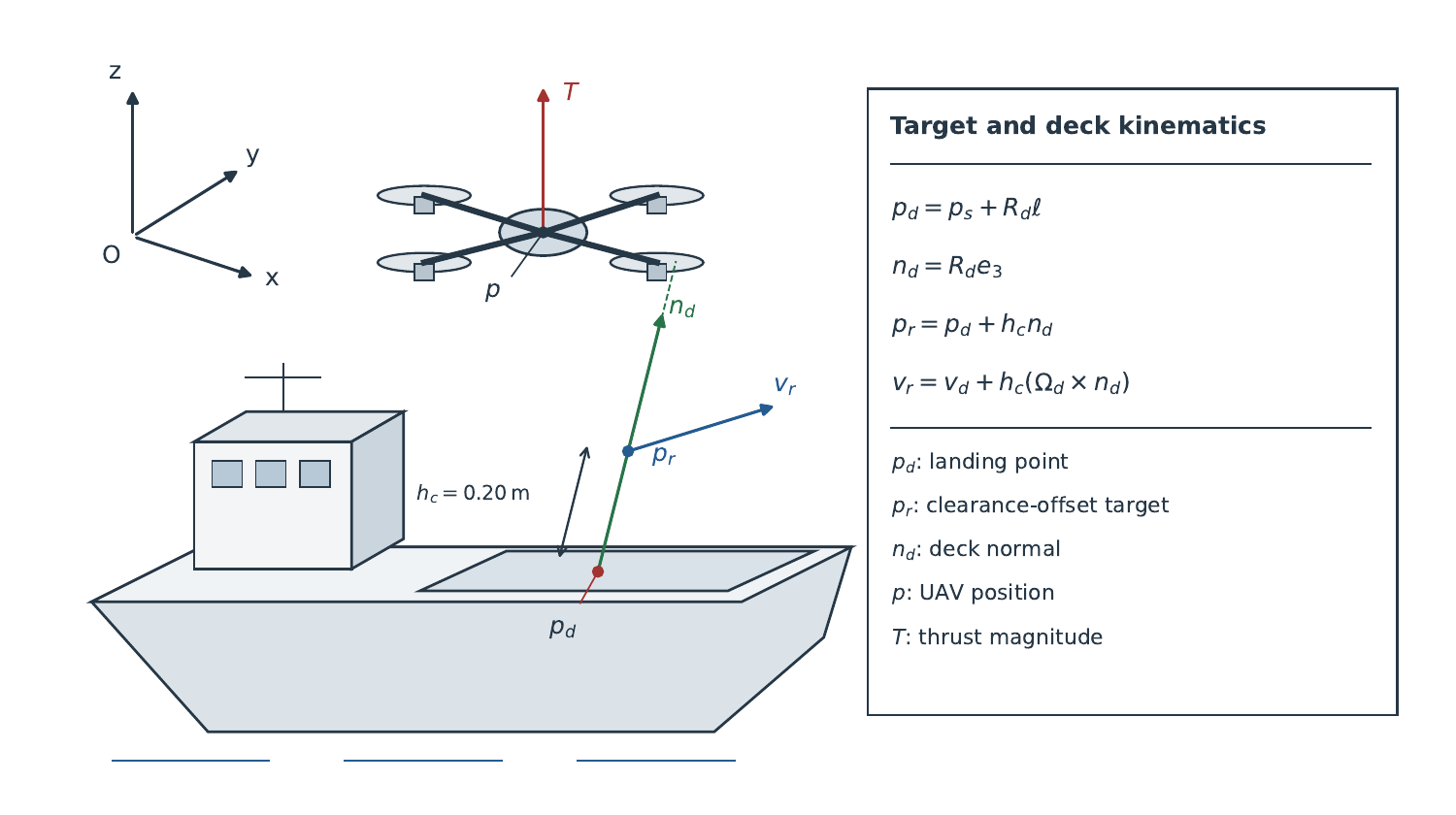}
\caption{Conceptual illustration of multirotor approach to a wave-excited offshore deck. The clearance-offset target is $h_c=0.20$~m above the landing point along the deck normal. Schematic constructed from geometric primitives for this study; the drawing source is provided with the manuscript. The vessel and multirotor are generic representations and are not drawn to scale. The schematic is not a calibrated vessel model or a physical touchdown demonstration.}
\label{fig:vessel}
\end{figure*}

\subsection{Prescribed sea motion and limitations}
The baseline ship-motion generator uses harmonic heave, roll and pitch. A separate irregular-wave branch constructs a deterministic finite-frequency realization with a JONSWAP-shaped spectrum and illustrative response filters. These are distinct model families and are not interchangeable with a calibrated vessel response model. The simulation does not include wind, motor response lag, estimator delay or aerodynamic interaction between ship and rotors. It also stops before physical deck contact: there is no impact, contact force, rebound, friction or landing-gear deformation. A compliant landing gear would compress and absorb energy during touchdown, but that process is outside this model. The terminal trajectory is therefore a simulated clearance-offset approach.

\subsection{Harmonic deck realization and reference kinematics}
In the baseline model, heave $h$, roll $\phi$ and pitch $\theta$ are prescribed as
\begin{align}
 h(t)&=A_h\sin(\omega_h t),\\
 \phi(t)&=A_\phi\sin(\omega_\phi t+\varphi_\phi),\\
 \theta(t)&=A_\theta\sin(\omega_\theta t+\varphi_\theta).
\end{align}
The baseline parameters are $A_h=0.40$~m, $\omega_h=0.50$~rad/s, $A_\phi=4^\circ$, $\omega_\phi=0.65$~rad/s, $\varphi_\phi=0.30$~rad, $A_\theta=3^\circ$, $\omega_\theta=0.55$~rad/s and $\varphi_\theta=-0.20$~rad. The landing-point offset in ship coordinates is $\ell=[2,0.5,0]^\top$~m. For ship reference position $p_s$, velocity $v_s$ and attitude $R_d$, the moving landing point and its velocity are
\begin{align}
 p_d&=p_s+R_d\ell,&v_d&=v_s+\Omega_d\times(R_d\ell),\\
 n_d&=R_de_3,&p_r&=p_d+h_cn_d,\\
 &&v_r&=v_d+h_c\Omega_d\times n_d.
\end{align}
The clearance target is displaced by $h_c=0.20$~m along the local deck normal $n_d$, not necessarily vertically upward in the inertial frame. It is a single offset target, not a hemispherical obstacle-free region. As the ship rotates, both the landing point and the clearance-offset target change position. Their velocities therefore depend on both the translational and rotational motion of the ship. The reference attitude and angular rate follow the prescribed deck orientation and its derivative.

\subsection{Deck angular velocity}
For the prescribed zero-yaw orientation $R_d=R_y(\theta)R_x(\phi)$, the angular velocity is
\begin{equation}
\boldsymbol\Omega_d^b=\begin{bmatrix}\dot\phi\\\dot\theta\cos\phi\\-\dot\theta\sin\phi\end{bmatrix},\qquad
\boldsymbol\Omega_d=R_d\boldsymbol\Omega_d^b.
\label{eq:v11_deckomega}
\end{equation}
The vector $\boldsymbol\Omega_d^b$ expresses the deck angular velocity in the deck-fixed coordinate system. Multiplication by $R_d$ expresses the same angular velocity in the inertial coordinate system, as required for calculating the landing-point and clearance-offset reference velocities above.

\subsection{Coordinate conventions and terminal error variables}
The inertial vertical unit vector is $\mathbf e_3=[0,0,1]^\top$, and the body $z$ axis defines the direction of thrust. The vehicle attitude is represented by a scalar-first unit quaternion satisfying $q^\top q=1$. The equivalent representations $q$ and $-q$ are treated consistently when evaluating attitude errors.

The position of the UAV relative to the landing point, expressed in deck-fixed coordinates, is $\mathbf r=R_d^\top(\mathbf p-\mathbf p_d)$. Its time derivative is
\begin{equation}
\dot{\mathbf r}=R_d^\top(\mathbf v-\mathbf v_d)-(R_d^\top\boldsymbol\Omega_d)\times\mathbf r.
\end{equation}
The second term accounts for the rotation of the deck coordinate system.

At the desired terminal condition, the UAV is located at a distance $h_c$ above the landing point along the deck normal. The desired deck-frame position and velocity are $\mathbf r=[0,0,h_c]^\top$ and $\dot{\mathbf r}=0$. The terminal errors are evaluated at the prescribed final time $T_F$, even when only the first part of the predicted control sequence is optimized.

The attitude error describes the difference between the UAV orientation and the deck orientation. It is represented locally by the rotation vector $\mathbf e_R=\operatorname{Log}(R_d^\top R(q))^\vee$. To evaluate angular-velocity tracking, the deck angular velocity is first expressed in the UAV body coordinate system. The corresponding error is $\mathbf e_\omega=\boldsymbol\omega-R(q)^\top\boldsymbol\Omega_d$. Both errors vanish when the UAV matches the orientation and angular velocity of the deck. These quantities are used in the tracking penalties. The rotation-logarithm representation is used locally, without assuming global smoothness.

\subsection{Deterministic irregular-wave construction}
The separate spectral branch evaluates a finite-frequency wave realization rather than a calibrated stochastic vessel response. In continuous notation, a representative JONSWAP-shaped elevation spectrum is
\begin{equation}
\begin{aligned}
S(\omega)&=\alpha g^2\omega^{-5}
\exp\!\left[-\tfrac54(\omega_p/\omega)^4\right]\\
&\quad\times\gamma^{\exp[-(\omega-\omega_p)^2/(2\sigma^2\omega_p^2)]},
\end{aligned}
\end{equation}
with peak frequency $\omega_p=2\pi/T_p$, shape factor $\gamma$, and a normalization chosen for the specified significant wave height. The finite realization uses $160$ frequencies and deterministic phases. Here $\eta_j(t)$ is the resulting deck-motion coordinate at time $t$, with $j\in\{h,\phi,\theta\}$ denoting heave, roll or pitch. The complex filter $H_j(\omega)$ specifies the amplitude and phase assigned to each wave-frequency component in that coordinate:
\begin{equation}
\begin{aligned}
\eta_j(t)=\sum_{i=1}^{160}&|H_j(\omega_i)|
\sqrt{2S(\omega_i)\Delta\omega_i}\\
&\times\cos(\omega_i t+\phi_i+\arg H_j(\omega_i)).
\end{aligned}
\end{equation}
The filters are illustrative, not measured response-amplitude operators for a named ship. Consequently, results cannot be interpreted as a vessel-specific sea-state certification. The time derivatives of the generated heave, roll and pitch signals give the deck translational and angular velocities. A point away from the ship reference origin also moves when the ship rotates; the cross-product term in the landing-point velocity accounts for this additional motion.

The original spectral implementation is more specific than a generic JONSWAP shape. With $z=\omega/\omega_p$ and $\sigma(\omega)=0.07$ for $\omega\leq\omega_p$ and $0.09$ otherwise, it uses
\begin{align}
S_0(\omega)&=g^2\omega^{-5}\exp[-1.25(\omega_p/\omega)^4]\gamma^{r(\omega)},\\
r(\omega)&=\exp\!\left[-\frac{(\omega/\omega_p-1)^2}{2\sigma(\omega)^2}\right],\\
H_h(\omega)&=\frac{0.75}{1+0.12z^2}e^{-i0.12z},\\
H_\phi(\omega)&=\frac{0.035z}{1+0.25z^2}e^{-i0.30z},\\
H_\theta(\omega)&=\frac{0.045z}{1+0.20z^2}e^{-i0.22z}.
\label{eq:v11_filters}
\end{align}
The spectrum is normalized so that $\sum_i S(\omega_i)\Delta\omega=H_s^2/16$, with $a_i=\sqrt{2S(\omega_i)\Delta\omega}$, and the original response and its derivative are
\begin{align}
\eta_j(t)&=\Re\sum_i H_j(\omega_i)a_i e^{i(\omega_i t+\varphi_i)},\\
\dot\eta_j(t)&=\Re\sum_i i\omega_iH_j(\omega_i)a_i e^{i(\omega_i t+\varphi_i)}.
\end{align}
These filters are illustrative, not vessel-identified response-amplitude operators.

\begin{figure*}[t]
\centering
\includegraphics[width=0.93\textwidth]{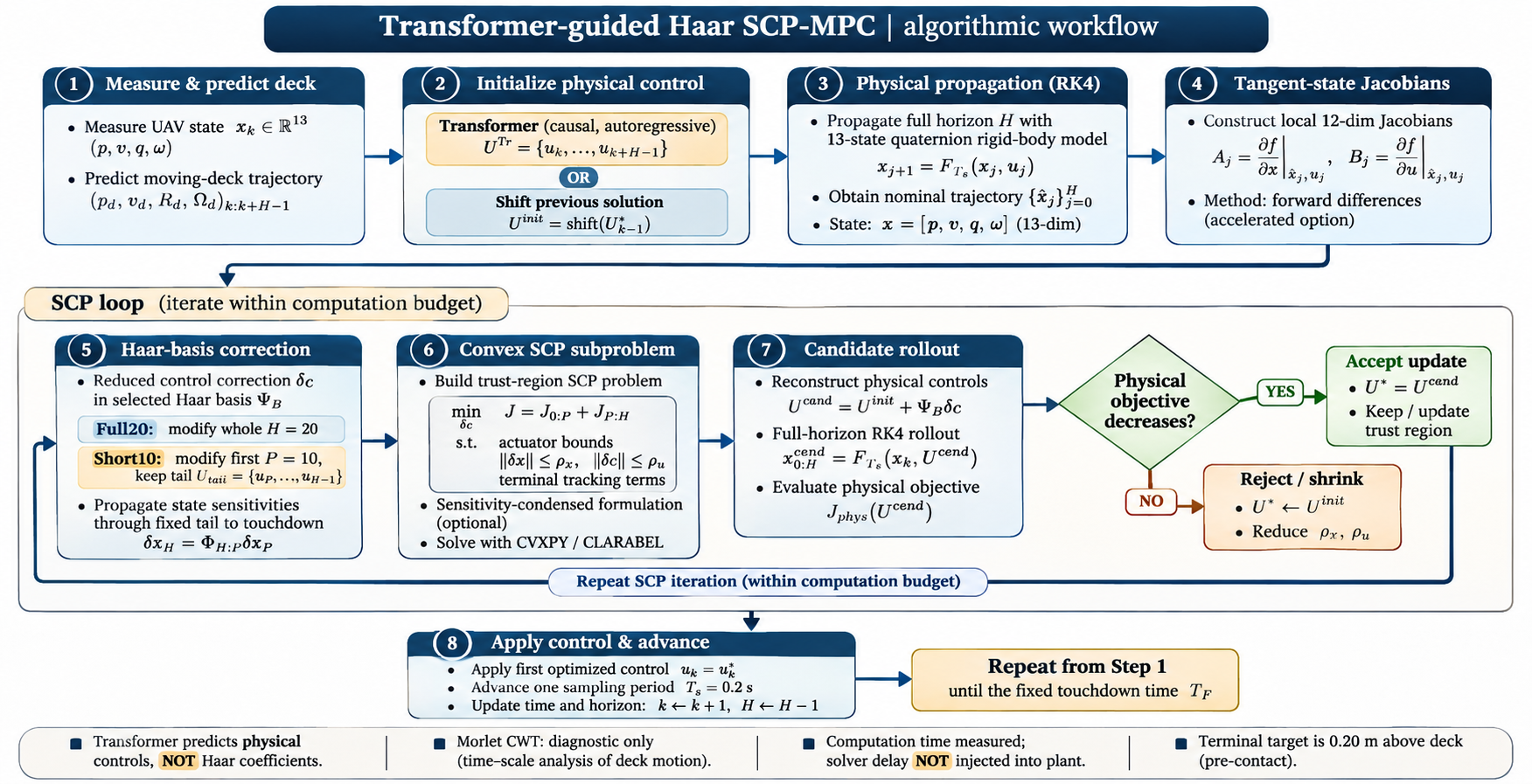}
\caption{Transformer-guided Haar SCP-MPC workflow. The earlier network proposes physical controls; HT10/FAA predicts Haar coefficients. SCP corrects the reconstructed controls, and RK4 rollout checks objective descent. Full20 and Short10 differ in the adjustable control segment; CL31--CL33 accelerate derivative and convex-problem calculations.}
\label{fig:workflow}
\end{figure*}

\section{Predictive control and sequential convex programming}
At each control update, MPC predicts the vehicle motion over a finite number $H$ of future sampling steps (the \emph{prediction horizon}). It compares that motion with a time-varying desired deck-relative trajectory (the \emph{reference}), while respecting thrust and torque limits (the \emph{constraints}). An optimizer selects a candidate control sequence that minimizes tracking errors and control effort. Only its first control is applied before the state is updated and the optimization is repeated; this is \emph{receding-horizon control}. Sequential convex programming (SCP) solves the trajectory problem through successive local convex approximations, checking each accepted candidate against the physical simulation model.

\begin{table}[t]
\centering
\caption{Controller specification for the numerical study. ``Available'' denotes ideal simulated information, not experimentally measured feedback.}
\label{tab:controller_spec}
\footnotesize
\setlength{\tabcolsep}{3pt}
\begin{tabular}{@{}p{.32\columnwidth}p{.61\columnwidth}@{}}
\toprule
Item & Implemented configuration \\
\midrule
Plant & Rigid multirotor, moving-deck pre-contact approach \\
Manipulated inputs & Total thrust $T$ and body torques $(\tau_x,\tau_y,\tau_z)$; not individual rotor commands \\
State supplied to MPC & Simulated position, velocity, quaternion attitude and body angular velocity (13 components); no sensor noise or estimator \\
Deck information & Prescribed deck position, velocity, orientation and angular rate, evaluated along the predicted horizon \\
Reference & Time-varying deck-relative approach, ending $0.20$ m along the deck normal above the landing point; reference velocity follows that offset target \\
Update / horizon & $T_s=0.20$ s; initially $H=20$ steps (4 s), then shrinking \\
Input constraints & Physical total-thrust and body-torque bounds on reconstructed controls; local SCP trust regions (see text) \\
Terminal assessment & Position error $\leq0.50$ m and relative-speed error $\leq0.30$ m/s; not touchdown certification \\
Solver & SCP convex subproblems using CVXPY/CLARABEL; RK4 physical rollout for acceptance \\
\bottomrule
\end{tabular}
\end{table}

At update $k$, the remaining horizon is $H=20-k$ for $k=0,\ldots,19$, with $T_s=0.2$~s and $T_F=4$~s. The predicted state satisfies $\mathbf x_{j+1}=F_{T_s}(\mathbf x_j,\mathbf u_j)$; references are evaluated at the corresponding absolute future times. The baseline physical objective is the special case $w_{R,j}=0$ of the following form; later FAA/PFAA variants add the stage-attitude weight $w_{R,j}$:
\begin{align}
J_{\rm phys}={}&\sum_{j=0}^{H-1}\bigl[
2\|\mathbf p_j-\mathbf p_{r,j}\|^2
+0.5\|\mathbf v_j-\mathbf v_{r,j}\|^2\nonumber\\
&\quad+0.01\|\mathbf u_j-\mathbf u_h\|^2
+w_{R,j}\|\mathbf e_{R,j}\|^2\bigr]\nonumber\\
&+100\|\mathbf p_H-\mathbf p_{r,H}\|^2
+20\|\mathbf v_H-\mathbf v_{r,H}\|^2\nonumber\\
&+20\|\mathbf e_{R,H}\|^2
+5\|\mathbf e_{\omega,H}\|^2.
\label{eq:cost}
\end{align}
For the FAA06 implementation, $w_{R,j}=w_{\rm stage}g_j$ with $w_{\rm stage}=1$ by default. The gain $g_j$ is evaluated on the current nominal trajectory and held fixed during each convex subproblem; it is recalculated following an accepted update. The original short-decision-horizon implementation (Short10 v2) has a different local objective, as specified below.

The SCP step linearizes the discrete dynamics around a nominal trajectory:
\begin{equation}
\begin{aligned}
\delta\mathbf x_{j+1}&=A_j\delta\mathbf x_j+B_j\delta\mathbf u_j,\\
A_j&=\partial F_{T_s}/\partial\mathbf x,\quad
B_j=\partial F_{T_s}/\partial\mathbf u.
\end{aligned}
\end{equation}
Local convex subproblems use input bounds and trust regions; proposed updates are checked by physical RK4 rollout and accepted only when the physical objective decreases by more than $10^{-7}$. The convex subproblems follow standard convex optimization principles \cite{boyd2004}. Sequential convex programming is discussed in \cite{kheirandishfard2021}; its convergence results do not establish global convergence for the present controller. The numerical optimization is implemented with CVXPY and CLARABEL, not a newly developed solver.

\subsection{Discrete MPC and local SCP formulation}
The original receding-horizon optimization is
\begin{equation}
\min_{U\in\mathcal U^H} J(U;x_k,t_k),\quad
x_{j+1}=F_{T_s}(x_j,u_j),\quad x_0=x_k.
\label{eq:v11_mpc_problem}
\end{equation}
At the nominal state and control, the RK4 map is linearized as
\begin{equation}
F_{T_s}(\bar x_j+\delta x_j,\bar u_j+\delta u_j)
\simeq F_{T_s}(\bar x_j,\bar u_j)+A_j\delta x_j+B_j\delta u_j,
\label{eq:v11_linearization}
\end{equation}
where the Jacobians are evaluated at the nominal trajectory,
\begin{equation}
A_j=\left.\frac{\partial F_{T_s}}{\partial x}\right|_{\bar x_j,\bar u_j},\qquad
B_j=\left.\frac{\partial F_{T_s}}{\partial u}\right|_{\bar x_j,\bar u_j}.
\label{eq:v11_jacobians}
\end{equation}
The original correction constraints are
\begin{align}
\delta x_0&=0,\\
\delta x_{j+1}&=A_j\delta x_j+B_j\delta u_j,\\
\|\delta x_j\|_2&\leq\rho_x,\quad\|\delta u_j\|_2\leq\rho_u,\\
u_{\min}&\leq\bar u_j+\delta u_j\leq u_{\max}.
\end{align}
With selected temporal basis $\Psi_B$ and a four-column coefficient matrix $C$, the physical input correction is
\begin{equation}
\delta U=\Psi_B C,\qquad C\in\mathbb R^{B\times4}.
\label{eq:v11_haar_correction}
\end{equation}
The SCP optimization variable in the Haar-parameterized formulation is the coefficient matrix $C$, not an independent second set of physical control corrections. The $B$ basis columns span the temporal input correction over the applicable decision segment, while the four columns of $C$ correspond to total thrust and the three body-frame torque channels. The fixed basis is not optimized. Each candidate is reconstructed as $\bar U+\Psi_B C$, and actuator limits and input trust regions are imposed on these \emph{physical} reconstructed commands. In the short-decision-horizon experiments, the correction basis acts on the first $P$ sampling intervals; later inputs are held fixed. This is distinct from the $20\times13$ basis used to decode the later Transformer's nominal proposal. The local state corrections use 12 tangent coordinates, including three for attitude, rather than treating the four quaternion components as independent Euclidean variables.

\subsection{Remaining-horizon optimization and acceptance test}
At a control update, the nominal input sequence is used to calculate a trajectory by RK4 integration through the remaining interval $[t_k,T_F]$. Local optimization searches for a correction to that sequence, rather than a new terminal time. The SCP subproblem uses first-order dynamics and a quadratic local tracking objective; the corrected input must obey actuator bounds and trust-region restrictions. Feasibility of the convex approximation alone does not imply feasibility of the trajectory calculated using the vehicle dynamics. Each candidate is therefore evaluated by a fresh RK4 rollout with the actual moving-deck references at the corresponding absolute times. The implemented acceptance rule tests \emph{descent}: $J_{\mathrm{phys}}(U_{\mathrm{candidate}})<J_{\mathrm{phys}}(U_{\mathrm{current}})-10^{-7}$. It is not an independent test of terminal landing feasibility. In particular, the reported $0.50$-m position and $0.30$-m/s relative-velocity thresholds are evaluated as outcome metrics, not separately enforced by this acceptance condition. Actuator bounds are imposed in the convex correction and the reconstructed inputs; no collision, impact, or landing-gear contact certificate follows from a decreasing physical objective. A candidate failing the descent test is rejected and the trust region is reduced. Only the accepted candidate determines the next nominal trajectory. The optimization is repeated with a newly measured simulated state after the first control input is applied.

The recorded wall time measures the timed controller section, which includes Jacobian construction, convex optimization and candidate RK4 rollouts where these operations occur inside that timed section. Some experiment wrappers measure the initial Transformer proposal separately; thus an SCP-update timing cannot be interpreted as the sum of Transformer inference and optimization unless both were included in the same timer. The archived results do not provide a consistently instrumented, across-variant breakdown of Transformer inference, Haar reconstruction, Jacobian assembly, convex solve and physical-rollout acceptance. No such component timings are inferred here. Computation time is not inserted into the simulated plant as a delay. Accordingly, a wall-time reduction and a closed-loop tracking result are separate observations, and neither alone establishes real-time implementability.

\subsection{Trust regions, acceptance, and computational accounting}
For a nominal trajectory $(\bar x_j,\bar u_j)$, local perturbations satisfy $\delta x_0=0$ and the linearized state recursion. The subproblem is constrained by physical actuator limits on $\bar u_j+\delta u_j$ and by Euclidean trust-region bounds $\|\delta x_j\|_2\leq\rho_x$ and $\|\delta u_j\|_2\leq\rho_u$, as specified in the baseline implementation. Haar coefficients themselves are not physical actuators and are not assigned the thrust and moment bounds directly. The reconstructed commands $\bar U+\Psi_B C$, not the coefficient matrix $C$, are subject to the physical input bounds and input trust-region restrictions. The state trust region applies to the 12-dimensional local tangent perturbation, not to four independent quaternion coordinates. The physical nonlinear rollout is recomputed for every candidate input sequence, and a candidate is accepted only when the evaluated full-horizon objective decreases by more than $10^{-7}$. In the baseline numerical implementation, the SCP solver has a default cap of eight outer iterations and permits up to five trust-region attempts per outer iteration, stopping if none is accepted. Later Short10 and matched closed-loop experiments use separately configured iteration caps; these settings must not be attributed to the original baseline solver. The accepted and rejected counts refer to physical-objective tests, not merely successful convex solver returns. Consequently, their wall-time differences are configuration comparisons rather than an isolated horizon-length effect.

For the original \texttt{mpc\_scp\_solve} implementation, the initial trust radii are $\rho_x=0.25$ in the local state coordinates and $\rho_u=2.0$ in the four control coordinates. On a rejected physical rollout, the radii are updated as
\begin{equation}
 \rho_x\leftarrow\max(0.01,0.5\rho_x),\qquad
 \rho_u\leftarrow\max(0.05,0.5\rho_u).
\end{equation}
After acceptance they are enlarged, subject to caps:
\begin{equation}
 \rho_x\leftarrow\min(0.5,1.15\rho_x),\qquad
 \rho_u\leftarrow\min(4.0,1.15\rho_u).
\end{equation}
These numerical limits describe the original baseline solver only. The Short10 and FAA/PFAA implementations also halve the radii after rejection, but do not apply the same minimum-radius clamps; their iteration caps and subsequent updates are variant-specific. Accordingly, the baseline rule must not be read as a universal rule for all experiments. Because state and input channels have different physical units, the reported Euclidean bounds are implementation-specific numerical choices rather than dimensionless geometric distances.

The implemented 12-dimensional tangent-state Jacobian is formed around a 13-component quaternion state by perturbing three local attitude coordinates rather than four unconstrained quaternion components. The original derivative implementation uses central differences, whereas the accelerated variant uses forward differences. If $z$ denotes a local tangent-state coordinate and $f(z,u)$ the corresponding discretized local map, the two approximations are
\begin{align}
\frac{\partial f}{\partial z_i}\bigg|_{\rm cen}&\simeq\frac{f(z+\varepsilon e_i,u)-f(z-\varepsilon e_i,u)}{2\varepsilon},\\
\frac{\partial f}{\partial z_i}\bigg|_{\rm fwd}&\simeq\frac{f(z+\varepsilon e_i,u)-f(z,u)}{\varepsilon}.
\end{align}
In the original baseline \texttt{local\_jacobians}, the finite-difference perturbation is the absolute value $\varepsilon=10^{-5}$ for each of the 12 tangent-state coordinates and four physical input channels. The matched forward-difference dynamics derivative calculation also uses $10^{-5}$, while its separate attitude-residual derivative calculation uses $10^{-6}$. These are coordinate-wise absolute steps, not automatically rescaled relative steps; their accuracy depends on the units and conditioning of each channel. Central differences use positive and negative perturbed RK4 evaluations, while forward differences reuse the nominal RK4 evaluation and one positive perturbation. The paired CL31--CL33 runs assess the resulting closed-loop differences and timing; they do not establish a general Jacobian error bound or numerical equivalence for all sea states.

\begin{figure}[t]
\centering
\fbox{\begin{minipage}{.91\columnwidth}\small
\centering
Measured/simulated UAV state and predicted deck trajectory\\[-1pt]
$\downarrow$\\[-1pt]
Transformer proposes a nominal control sequence\\[-1pt]
$\downarrow$\\[-1pt]
Physical rollout and local dynamics derivatives\\[-1pt]
$\downarrow$\\[-1pt]
Haar-coefficient SCP correction ($P$ adjustable steps)\\[-1pt]
$\downarrow$\\[-1pt]
Full $H$-step physical rollout and cost acceptance\\[-1pt]
$\downarrow$\\[-1pt]
Apply first command; advance state; repeat
\end{minipage}}
\caption{Receding-horizon workflow. The Transformer supplies an initial proposal; it does not replace the physics model or the convex correction solver. FAA/PFAA modify attitude weights in the correction objective. CL31--CL33 change derivative assembly and/or the size of the convex subproblem, not the control objective.}
\label{fig:controller_steps}
\end{figure}

\section{Transformer initialization and Haar control coordinates}
The learned sequence model is trained on expert control trajectories and generates a nominal physical thrust--torque sequence $\widehat U_{\rm Tr}$. Its conditioning includes the physical state, task goal, time and return-/constraint-to-go quantities; the proposal is physically propagated before SCP correction. The learned model is an initializer, not a replacement for the rigid-body plant.

For a corrected horizon of length $P$, the control correction is restricted to a selected orthonormal Haar basis $\Psi\in\mathbb R^{P\times B}$:
\begin{equation}
\delta U_{0:P-1}=\Psi\,\delta C,\qquad
\delta\mathbf c=\mathrm{vec}(\delta C)\in\mathbb R^{4B}.
\end{equation}
In the full-horizon comparison, $N=20$ and $B=13$ (52 correction scalars rather than 80); in the short-horizon implementation, $P=10$ and $B=8$ (32 correction scalars). As the landing horizon shrinks, the basis is reduced consistently. The original nominal control tail remains fixed beyond the corrected segment. A Morlet continuous wavelet transform is used only as a diagnostic of deck-reference variation, not as a replacement plant model or active prediction layer.

\subsection{Non-dyadic Haar construction and deck-informed selection}
For a horizon of length $M$, let $\Psi\in\mathbb R^{M\times M}$ be a discrete orthonormal Haar-like matrix with $\Psi^\top\Psi=I_M$ and constant column $\psi_0=M^{-1/2}\mathbf 1$. The construction accommodates the non-power-of-two initial horizon $M=20$. With selected index set $\mathcal I_B$, $\Psi_B=\Psi[:,\mathcal I_B]$, the correction is $\delta U=\Psi_B C$. For a recursively split interval $I=I_L\cup I_R$ with $m=|I|$, $m_L=|I_L|$ and $m_R=|I_R|$, the original normalized detail vector is
\begin{equation}
\psi_I(i)=\begin{cases}
\sqrt{m_R/(mm_L)},&i\in I_L,\\
-\sqrt{m_L/(mm_R)},&i\in I_R,\\
0,&\text{otherwise}.
\end{cases}\label{eq:v11_haarsplit}
\end{equation}
Together with $\psi_0=M^{-1/2}\mathbf1$, recursively splitting intervals gives an orthonormal basis even when $M$ is not a power of two. The fixed-basis variant selects a predetermined coarse-to-fine subset. The deck-informed variant projects normalized deck-reference features $Z\in\mathbb R^{M\times d}$ onto the basis, $D=\Psi^\top Z$, and ranks the detail columns by
\begin{equation}
E_i=\sum_{\ell=1}^d D_{i\ell}^2,\qquad
\mathcal I_B=\{0\}\cup\operatorname{Top}_{B-1}\{E_i:i\ge1\}.
\end{equation}
Retaining the constant column guarantees access to a uniform correction direction. The energy ranking minimizes a reference-feature projection error among the chosen orthonormal columns; it does not minimize the physical MPC objective. The initial full-horizon control correction has $4B=52$ scalar coefficients at $B=13$ rather than $4M=80$ unconstrained scalar corrections. This count excludes state variables, constraints and rollout operations and is not a claim of proportional optimization speedup.

A two-resolution move-blocked predecessor used six independent early intervals followed by seven two-interval blocks. In a separate diagnostic, a complex Morlet transform of sampled deck features was computed over 48 candidate periods spanning 0.4--12 s:
\begin{align}
W_z(a,b)&=a^{-1/2}\int z(t)\psi^*((t-b)/a)\,dt,\\
\psi(\xi)&\propto e^{i\omega_0\xi}e^{-\xi^2/2}.
\end{align}
The transform provided time-scale diagnostics; it neither replaced the RK4 plant nor produced an independently validated adaptive MPC grid.

\subsection{Different roles of Haar parameterization and Transformer initialization}
Move blocking reduces the number of optimization variables by keeping the control input constant over selected time intervals. Haar parameterization provides an alternative representation in which broad control adjustments and finer local corrections can be combined within the same optimization problem. This allows the controller to represent control variations at different time scales using a limited number of coefficients. Although the selected Haar basis is orthonormal, the physical vehicle dynamics remain unchanged. The dynamics are linearized around the nominal trajectory, and the Jacobians are recalculated after each accepted SCP step.

The Transformer serves a different purpose. The original autoregressive network learns to predict physical control sequences, whereas the later HT10/FAA network predicts coefficients representing expert control sequences. In both cases, the predicted sequence provides an initial solution for the subsequent physics-based optimization. The Transformer does not determine the SCP correction, select the correction basis, or guarantee that the terminal requirements will be satisfied.

Even a predicted control sequence with small average prediction errors may lead to considerable terminal tracking errors, particularly in velocity. It is therefore necessary to evaluate the proposed control sequence using the physical vehicle model. The Transformer initialization should also be compared with the conventional approach of shifting the previous optimal control sequence forward by one time step. This conventional approach is computationally inexpensive and can provide an accurate initial solution when the vehicle state and deck motion change gradually.

\subsection{Time representation and sequential Transformer prediction}
The original network uses a learned absolute-timestep embedding shared across token modalities. For modality $m$ with token features $\xi_k^{(m)}$,
\begin{equation}
z_k^{(m)}=E_m(\xi_k^{(m)})+E_t(k).
\label{eq:v11embedding}
\end{equation}
The time-to-go is implicit in the absolute index for the fixed landing time:
\begin{equation}
t_k=kT_s,\qquad \tau_{\mathrm{go},k}=T_F-kT_s=(N-k)T_s.
\label{eq:v11time}
\end{equation}
This does not demonstrate the usefulness of a separate time-to-go token. The 28 expert trajectories were divided into 22 training and six held-out trajectories. The recorded fine-tuning used 60 epochs, AdamW learning rate $10^{-5}$ and weight decay $10^{-4}$, with the original checkpoint normalization retained. Mean squared error of normalized control predictions evaluated using the reference training sequence changed from 0.7754 to 0.7042; this is not a closed-loop performance measure.

For the recorded control-effort reward $r_j=-\tfrac12\|u_j-u_h\|_2^2$, the return-to-go obeys
\begin{equation}
G_j=\sum_{i=j}^{N-1}r_i,\quad
G_{j+1}=G_j+\tfrac12\|u_j-u_h\|_2^2.
\label{eq:v11rtg}
\end{equation}
Let $\mathcal H_{j|k}$ contain the causally available normalized state and previous-action history, deck-goal tokens, RTG, CTG and absolute timesteps. The autoregressive proposal is
\begin{align}
\widetilde u_{j|k}&=f_\theta(\mathcal H_{j|k},g_j,G_j,C_j),\label{eq:v11prediction}\\
\widehat u_{j|k}&=\Pi_{\mathcal U}
 \bigl(\mu_u+\sigma_u\odot\widetilde u_{j|k}\bigr),\\
\widehat x_{j+1|k}&=F_{T_s}(\widehat x_{j|k},\widehat u_{j|k}),
\quad\widehat x_{k|k}=x_k,\label{eq:v11rollout}\\
\mathcal H_{j+1|k}&=\mathcal A(\mathcal H_{j|k},\widehat u_{j|k},
 \widehat x_{j+1|k},g_{j+1},G_{j+1},C_{j+1}).\label{eq:v11history}
\end{align}
Here $\mathcal A$ appends the new history tokens and $\Pi_{\mathcal U}$ clips to the physical actuator box and the inference routine supplies CTG as zero-valued conditioning; it does not compute verified future constraint violations. Absolute timestep indices are not reset at every MPC update. The proposed controls initialize, but do not replace, the physics-based SCP solver. The local sensitivity mechanism for proposal error is
\begin{equation}
\delta x_{j+1}\simeq A_j\delta x_j+B_j\delta u_j.
\label{eq:v11proposalerror}
\end{equation}
This relation does not by itself prove that autoregressive error accumulation caused the observed large proposal-only rollout errors.

\subsection{Expert demonstrations, conditioning and inference}
The original expert dataset contained 28 physics-based SCP trajectories, divided into 22 training and six held-out trajectories. A trajectory is a sequence of simulated states and optimized actions, not a physical flight demonstration. The causal Transformer conditions its autoregressive physical-action prediction on normalized state and previous-action tokens, future deck-goal information, return-to-go (RTG), constraint-to-go (CTG), and absolute timestep embeddings. Causal masking excludes unavailable future action tokens. At an update, the generated thrust--moment sequence is clipped or otherwise checked against physical input bounds and propagated by the original RK4 model to form a nominal state trajectory. SCP then optimizes an input correction around that trajectory.

A horizon-aware RTG initialization diagnostic did not outperform the original initialization: mean thrust RMSE relative to expert controls was $1.6097$ N for the horizon-aware variant versus $1.4782$ N for the original. An independent physical-rollout diagnostic exposed large early-horizon proposal errors, so action-token prediction accuracy alone is not a surrogate for terminal feasibility. Neither the Transformer nor its training objective replaces the physical rollout or enforces the terminal velocity constraint by construction.

\subsection{Distinct HT10/FAA coefficient-prediction implementation}
The later matched HT10 and FAA experiments do \emph{not} invoke the preceding autoregressive physical-action model. Their notebook checkpoint defines a two-layer, four-head causal Transformer encoder with an input embedding from 28 features to 64 hidden dimensions and a 52-dimensional output head reshaped into a $13\times4$ coefficient array. The 28 input features at each of the 20 nominal time indices consist of a 13-component state, a 13-component deck-goal state, absolute time, and remaining time. In the recorded evaluation wrapper the state feature is the initial state repeated at every token position; this is a fixed-context proposal, not a simulated autoregressive state rollout. The goal features populate deck-reference position and velocity, with other goal components set to zero. The remaining-time feature is $19T_s-t_j$ in the saved wrapper, rather than $T_F-t_j$.

For a frozen $20\times13$ Haar matrix $B_{20}$, network output $C_\theta\in\mathbb R^{13\times4}$, checkpoint output scale $s_y$, and hover input $U_h$, the actual proposal is
\begin{equation}
U_{\mathrm{proposal}}=\Pi_{[u_{\min},u_{\max}]}\bigl(U_h+B_{20}(s_y C_\theta)\bigr),
\label{eq:ht10decode}
\end{equation}
where clipping is componentwise. The benchmark then uses a separate short-horizon correction basis $B_{10}\in\mathbb R^{10\times8}$, obtained from the rank-reduced QR/SVD orthogonalization of the first ten rows of $B_{20}$. Thus the \emph{proposal} coefficient representation and the \emph{correction} coefficient representation are different. The HT10/FAA evaluation does not use the RTG/CTG token recursion of the earlier autoregressive model. This separation is essential when interpreting the reported training-size and FAA/PFAA experiments. In particular, the Haar basis has two implemented roles: it decodes the later network's coefficient-valued proposal and it parameterizes the subsequent SCP correction. The earlier autoregressive network predicts physical controls and does not use that same coefficient-output head. The present experiments do not evaluate wavelet-scale deck-motion features as Transformer inputs, nor do they establish an adaptive-horizon method driven by those features. Such designs are possible future comparisons, not part of the controller assessed here.

The matched 30-scenario attitude-weighting comparison (FAA06) reuses one frozen Transformer proposal for all three modes within each scenario, then calls \texttt{short\_horizon\_scp\_faa\_v1} with \texttt{mode=fixed/faa/pfaa} and \texttt{max\_iterations=3}. Consequently, the three-way comparison isolates the mode configuration under a common proposal for each scenario, but does not isolate the effect of Transformer architecture.

\section{Terminal-aware short-decision-horizon SCP}
The reduced-decision-horizon controller modifies only the first part of the planned control sequence. The remaining controls are kept at their nominal values. Nevertheless, changes to the initial controls influence the vehicle throughout the entire prediction interval. The controller must therefore account for their effects at the prescribed final landing time, rather than evaluating performance only where the optimized segment ends.

Let $H$ denote the full number of future sampling steps and $P<H$ the number of steps whose controls may be changed by the optimizer. In the main short-horizon experiment, $H=20$ and $P=10$. The first $P$ controls are adjusted; the later $H-P$ controls remain fixed. If
$\delta\mathbf u_j=D_j\delta\mathbf c$ for $j<P$ and $\delta\mathbf u_j=0$ thereafter, then
\begin{equation}
\begin{aligned}
S_0&=0,\\
S_{j+1}&=A_jS_j+B_jD_j,\\
\delta\mathbf x_j&=S_j\delta\mathbf c,\quad j=0,\ldots,H.
\end{aligned}
\label{eq:sensitivity}
\end{equation}
This effect is calculated using the state-transition matrices of the locally linearized dynamics. The matrix $\Phi_{H:P}=A_{H-1}\cdots A_P$ propagates a small state change from the end of the optimized segment, at step $P$, to the final prediction step $H$. Multiplying this matrix by the sensitivity $S_P$ gives the terminal sensitivity $S_H=\Phi_{H:P}S_P$. The terminal position and velocity penalties can then be related to the controls being optimized. Attitude and angular-rate errors require additional residual sensitivities because orientation is represented using rotations rather than ordinary Cartesian coordinates.

\emph{Implementation distinction:} The earlier short-decision-horizon convex subproblem includes terminal position and velocity penalties but not explicit terminal attitude or angular-rate penalties. Those additional errors are evaluated in the full physical acceptance objective. The later FAA/PFAA variants introduce attitude alignment in the local optimization, as described below. Thus the preceding sensitivity identity explains the terminal transport; it does not claim that all terminal penalties appear in every convex subproblem.

\subsection{Sensitivity of terminal position, speed and orientation}
To see how an early control correction affects the final state, define the state-transition product
\begin{equation}
 \Phi_{j:i}=A_{j-1}A_{j-2}\cdots A_i,\qquad j>i,
\end{equation}
with $\Phi_{i:i}=I$. For example, $\Phi_{6:3}=A_5A_4A_3$: the earliest one-step matrix acts first and is therefore on the right. This product maps a small state change at step $i$ to its effect at step $j$. After step $P$, the input corrections are zero, but the effect of the earlier corrections on the state remains: $S_j=\Phi_{j:P}S_P$ for $j>P$. This product is also needed for each intermediate step of the fixed-control tail. For rotational tracking, the terminal residual Jacobian must be evaluated in the local three-dimensional attitude coordinates rather than by treating all four quaternion components as independent variables. The exact penalties present in each convex subproblem are stated in the following subsections.

\subsection{Short-decision-horizon local objective}
In the original notation, $H=20$, $P=10$, and the correction is restricted to the first $P$ controls:
\begin{align}
\delta x_{j+1}&=A_j\delta x_j+B_j\delta u_j,\quad\delta x_0=0,\\
\delta U_{0:P-1}&=\Psi_P\delta C,\quad
\delta u_j=0\quad(P\leq j<H),\\
\delta x_H&=\Phi_{H:P}\delta x_P,\quad
\Phi_{H:P}=A_{H-1}\cdots A_P.
\end{align}
The complete \emph{original} local objective is
\begin{equation}
\begin{aligned}
\min_{\delta C}\quad&
\sum_{j=0}^{H-1}\bigl[
 2\|\bar p_j+\delta p_j-p_{r,j}\|_2^2\\[-2pt]
&\qquad+0.5\|\bar v_j+\delta v_j-v_{r,j}\|_2^2\\[-2pt]
&\qquad+0.01\|\bar u_j+\delta u_j-u_h\|_2^2\bigr]\\
&+100\|\bar p_H+\delta p_H-p_{r,H}\|_2^2\\
&+20\|\bar v_H+\delta v_H-v_{r,H}\|_2^2 .
\end{aligned}
\label{eq:v11shortobjective}
\end{equation}
Although only the first $P$ controls are adjustable, the stage tracking penalties are evaluated over all $H$ steps. The later fixed controls still affect the predicted path, and the effect of the early corrections is propagated through those steps. An important distinction is that this earlier convex objective contains no explicit terminal attitude or angular-rate terms. The full nonlinear physical acceptance objective evaluates those additional terms, so local optimization and acceptance are not the same objective. FAA/PFAA extensions are separate audited variants, not retrospective descriptions of this original subproblem.

\subsection{Terminal prediction using the remaining fixed control sequence}
The first $P$ planned commands are adjusted by the optimizer; the later commands remain at the nominal values. The UAV still responds to those early adjustments during the remaining $H-P$ steps. In symbols, the corrected input segment is $j=0,\ldots,P-1$; the tail $j=P,\ldots,H-1$ remains fixed at the nominal proposal. With $\Phi_{H:P}=A_{H-1}\cdots A_P$, the terminal state perturbation is $\delta x_H=\Phi_{H:P}\delta x_P$. More generally, each stage of the fixed tail has $\delta x_j=\Phi_{j:P}\delta x_P$, so the tail contributes to the linearized physical tracking cost even though its control variables are fixed. A terminal quadratic approximation has the schematic form
\begin{equation}
\ell_H\simeq \|e_{p,H}+E_p S_H\delta c\|_{Q_p}^2+
\|e_{v,H}+E_v S_H\delta c\|_{Q_v}^2,
\end{equation}
with additional attitude and rate terms where included in the specific audited subproblem. The original short-decision-horizon implementation (Short10 v2) convex objective included terminal position and velocity terms but \emph{did not include} terminal attitude or angular-rate terms, although the physical acceptance objective evaluated these quantities. Its original running cost includes position, velocity and control effort throughout the remaining $H$ steps, including the fixed-input tail; see Eq.~\eqref{eq:v11shortobjective}. Later FAA/PFAA variants add explicit alignment terms to the convex objective. These formulations are not retrospectively described as identical.

\enlargethispage{\baselineskip}
\subsection{Convex optimization formulation for FAA/PFAA}
The FAA/PFAA implementation defines the landing-point deck normal $n_j$, landing-point position $p_{d,j}$ and velocity $v_{d,j}$. The deck-relative height and closing speed are
\begin{equation}
 h_j=n_j^\top(p_j-p_{d,j}),\qquad
 c_j=-n_j^\top(v_j-v_{d,j}).
\end{equation}
The controller uses three ways to weight attitude tracking during the approach. The \emph{fixed} mode uses the same stage weight at every time step. The height-dependent mode (FAA) increases the weight as the vehicle approaches the deck along its normal. The predicted-arrival mode (PFAA) increases the weight as the estimated remaining closing time becomes small, but falls back to the fixed weight when closing speed is too low. Mathematically, the three gains are
\begin{equation}
\begin{aligned}
 g_j^{\rm fixed}&=1,\\
 g_j^{\rm FAA}&=1+\alpha\exp[-\beta\max(h_j,0)],\\
 g_j^{\rm PFAA}&=\begin{cases}
 1,&c_j\leq c_{\min},\\
 1+\alpha\exp[-\tau_j/\tau_a],&c_j>c_{\min},
 \end{cases}\\
 \tau_j&=\max(h_j,0)/c_j\quad(c_j>c_{\min}).
\end{aligned}
\label{eq:faa_gain_verified}
\end{equation}
The defaults are $\alpha=3$, $\beta=2$, $\tau_a=1$~s and $c_{\min}=0.02$~m/s. Here $h_j$ is signed height along the instantaneous deck normal, $c_j$ is the approach speed along that normal (positive when closing), and $\tau_j$ is a simple height-over-speed estimate of time remaining; it is not a simulated touchdown time. FAA and PFAA change the stage attitude penalty, not the terminal feasibility thresholds. With $e_{R,j}$ the three-component deck-relative attitude error and $G_{R,j}$ its central-difference Jacobian in the right-multiplicative 12-state tangent coordinates, the FAA local objective is
\begin{equation}
\begin{aligned}
\min_{\delta C}\;&\sum_{j=0}^{H-1}\bigl[
2\|\bar p_j+\delta p_j-p_{r,j}\|^2\\
&+0.5\|\bar v_j+\delta v_j-v_{r,j}\|^2
+0.01\|\bar u_j+\delta u_j-u_h\|^2\\
&+w_{\rm stage}g_j\|e_{R,j}+G_{R,j}\delta x_j\|^2\bigr]\\
&+100\|\bar p_H+\delta p_H-p_{r,H}\|^2\\
&+20\|\bar v_H+\delta v_H-v_{r,H}\|^2\\
&+20\|e_{R,H}+G_{R,H}\delta x_H\|^2\\
&+5\|e_{\omega,H}+G_{\omega,H}\delta x_H\|^2.
\end{aligned}
\label{eq:faa_objective_verified}
\end{equation}
The correction satisfies $\delta u_{0:P-1}=\Psi_P\delta C$ and $\delta u_{P:H-1}=0$. Dynamics sensitivities are propagated for all $H$ steps, including the tail. The implementation imposes $\|\delta x_j\|_2\leq\rho_x$, $\|\delta u_j\|_2\leq\rho_u$ and actuator bounds for $j<P$, plus $\|\delta x_P\|_2\leq\rho_x$; it does not impose those trust-region inequalities at every later tail state. The gain $g_j$ is a fixed coefficient within each convex solve, not an optimization variable. For the PFAA branch, the $c_j\leq0.02$~m/s condition returns the baseline gain even when the vehicle is moving away from the deck; the time-to-contact proxy is therefore used only for positive closing speeds above this threshold. Each outer iteration permits up to five trust-region attempts and accepts only a physical-rollout cost reduction greater than $10^{-7}$. The default cap is three outer iterations. All three modes use identical terminal weights $(100,20,20,5)$; only $g_j$ differs. The feasibility flag tests $\|e_{p,H}\|\leq0.50$~m and $\|e_{v,H}\|\leq0.30$~m/s and does not certify attitude or touchdown.

FAA and PFAA are stage-weighting strategies for the existing deck-relative attitude objective, not newly introduced deck-attitude synchronization. Their comparison is meaningful only with the same 30 sea-motion scenarios, action initialization, corrected horizon and acceptance protocol. The terminal alignment improvement is reported separately from terminal position, velocity and feasibility.

\subsection{Sensitivity-condensed SCP}
The uncondensed SCP problem includes corrections to the state at every future time, even though those state corrections are determined by the input corrections and the local dynamics. Equation~\eqref{eq:sensitivity} expresses each state correction directly in terms of the Haar coefficient changes. Substituting these expressions removes the redundant state-decision variables from the convex subproblem without changing the nominal vehicle simulation. Stacking the weighted local residuals gives
\begingroup
\setlength{\belowdisplayskip}{3pt}
\setlength{\belowdisplayshortskip}{3pt}
\begin{equation}
\min_{\delta\mathbf c}\;\|M\delta\mathbf c+b\|_2^2
\label{eq:condensed}
\end{equation}
\endgroup
subject to input and trust-region bounds, where $M$ contains the condensed state, attitude and control sensitivities. This is a \emph{condensed formulation} of an SCP subproblem, not a novel convex solver. For the shrinking horizon, $H$ and the basis dimensions change at each update, so the condensed problem is rebuilt with consistent dimensions. The physical rollout, trust-region acceptance and fixed-tail propagation remain unchanged.

To construct the convex subproblem, SCP needs the local response of the simulated vehicle to small changes in its state and commands. The local Jacobians describe how a small change in the current state or control changes the state predicted one sampling step later. They are estimated numerically by perturbing each coordinate and repeating the model step. Central differences compare predictions with a small positive and negative perturbation; forward differences compare the positive perturbation with the unperturbed prediction. The state perturbations use 12 local coordinates, including a three-dimensional orientation perturbation, instead of varying the four quaternion components independently. The baseline Jacobians use central finite differences. Forward differences require fewer model evaluations, at the possible cost of a less accurate local derivative. These are independent modifications: condensation reduces the local optimization representation, while forward differences accelerate derivative assembly. Their combined effect is measured rather than estimated by multiplying separate speedup factors.

\section{Numerical experiments and results}
For traceability, we retain concise identifiers for the main experiment families. HT10 is the 30-scenario comparison between optimizing all 20 future controls and optimizing only the first 10, with the remaining 10 fixed. FAA05 is the four-scenario pilot for fixed, height-dependent (FAA) and closing-time-dependent (PFAA) attitude weighting; FAA06 extends this comparison to 30 scenarios with a common Transformer proposal within each scenario. CL31 compares central and forward finite-difference derivatives, CL32 compares the original and sensitivity-condensed optimization formulations, and CL33 combines both changes in a matched 20-update closed-loop run. These identifiers are retained here only after this definition so that readers can trace the results to the code. The project contains several benchmark families that should not be pooled: (i) first-update planning over 30 sea-motion scenarios; (ii) separate 20-update closed-loop Haar and Transformer experiments; (iii) four- and 30-scenario FAA/PFAA comparisons; (iv) training-size scaling; and (v) matched 20-update derivative/condensation acceleration. First-update feasibility is a prediction criterion, not an observed physical landing.

\begin{table}[t]
\centering\small
\caption{What changes between the main experiment families.}
\label{tab:methods_guide}
\begin{tabular}{@{}p{.21\columnwidth}p{.71\columnwidth}@{}}
\toprule
Label & Scientific comparison\\
\midrule
HT10 & Full 20-step versus first-10-step control correction, with full terminal propagation\\
FAA/PFAA & Fixed versus height-based versus closing-time-based stage attitude weights\\
CL31 & Central versus forward finite-difference Jacobians\\
CL32 & State-and-control versus sensitivity-condensed convex subproblem\\
CL33 & Forward differences and condensation together, under matched closed-loop conditions\\
\bottomrule
\end{tabular}
\end{table}

\subsection{Types of numerical experiments and terminal tracking criteria}
Three separate experimental levels are used. An initial-update test solves a planning problem from a specified state and compares predicted terminal errors. A receding-horizon simulation executes successive planned first inputs and propagates the idealized rigid-body plant; it does not account for the elapsed optimization wall time. A component audit tests numerical equivalence or timing for a derivative calculation or convex subproblem, but does not by itself establish closed-loop behavior. This distinction prevents combining first-update feasibility counts with closed-loop final errors or component-only speedups.

The first-update feasibility indicator is a satisfaction of both terminal error thresholds, not a scalar objective value. The evaluation protocol reports position, speed, attitude, and angular-rate errors as separate terminal measures. The additional FAA/PFAA runs evaluate the latter quantities more explicitly, but still do not simulate contact. Across all families, reported time is measured on the experimental computing environment; results may vary with runtime load, solver configuration and warm-up effects.

\subsection{Full- and short-decision-horizon first-update comparison}
In the HT10 matched 30-scenario first-update benchmark, full-horizon Haar SCP satisfied the combined $0.50$-m position and $0.30$-m/s velocity criteria in 29/30 scenarios; terminal-aware Short10 satisfied them in 13/30. Short10 met the position threshold in all 30 cases, but failed the velocity criterion in 17. Median measured wall time decreased from 8.84 to 6.12~s. The faster short-decision-horizon configuration therefore did \emph{not} preserve the full-horizon feasibility rate.

\begin{figure*}[t]
\centering
\includegraphics[width=.86\textwidth]{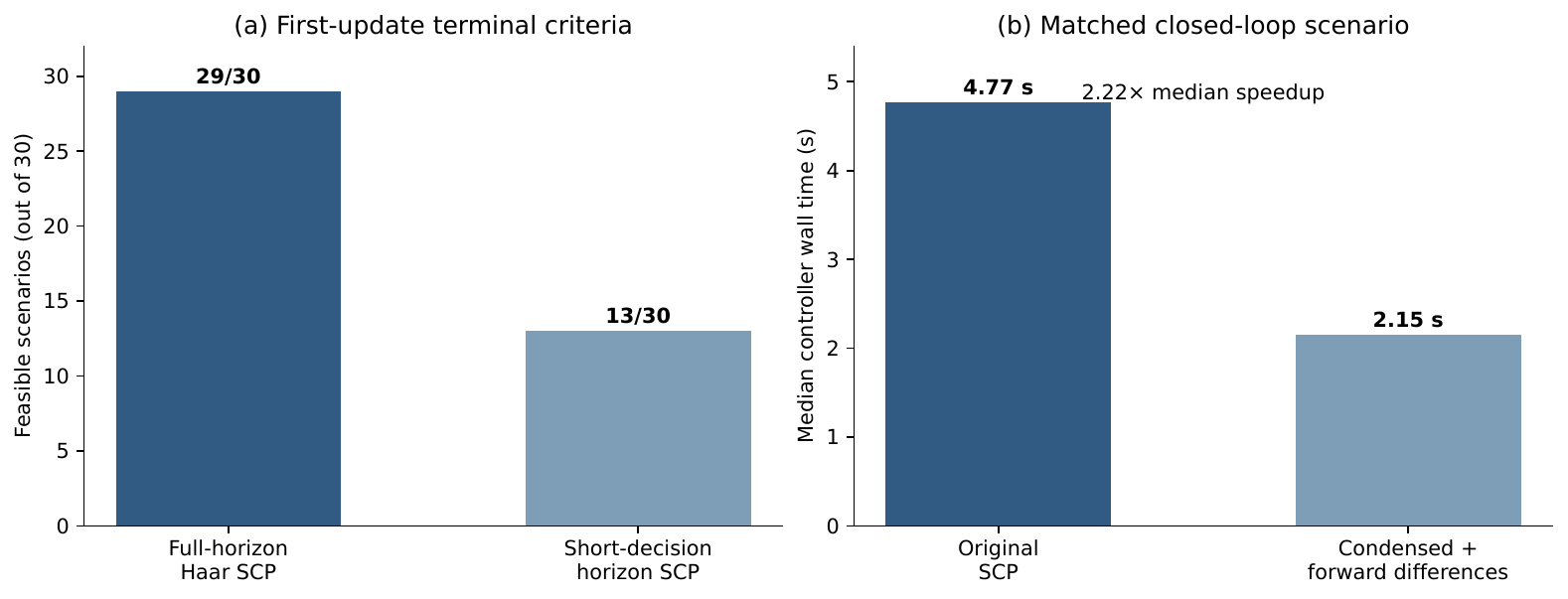}
\caption{Two separate archived comparisons. Left: first-update predicted feasibility for full- and short-decision-horizon control over 30 scenarios. Right: median computation time for the matched single-scenario 20-update acceleration comparison. These panels use different protocols and must not be interpreted as a common statistical experiment.}
\label{fig:results_comparison}
\end{figure*}

\subsection{Training-size and attitude-alignment comparisons}
\begin{table*}[t]
\centering\caption{Matched 30-scenario first-update Short10 comparisons by Transformer training size. The RMSE is from a separate held-out prediction evaluation.}
\begin{tabular}{lrrr}\toprule
& T72 & T400 & T2000\\\midrule
Training examples &72&400&2000\\
Held-out thrust RMSE (N)&0.030298&0.030074&0.011715\\
Median terminal position error (m)&0.02491&0.03190&0.02544\\
Median terminal speed error (m/s)&0.31419&0.31782&0.31338\\
Predicted feasible cases&13/30&12/30&12/30\\
Median total wall time (s)&4.366&4.744&5.050\\\bottomrule
\end{tabular}
\end{table*}
Increasing training size substantially improved held-out thrust prediction but did not increase predicted short-horizon feasibility. This does not prove that additional or more diverse training data cannot help; it shows that supervised action accuracy alone was not sufficient in the tested configurations.

In the separate 30-scenario FAA06 comparison, fixed, FAA and PFAA weighting each yielded 13/30 predicted feasible cases. Mean terminal position errors were 0.05862, 0.05855 and 0.05770~m, respectively, while mean attitude errors were 2.0767, 2.0668 and 2.0761 degrees. PFAA modestly improved mean position error, but did not improve feasibility; FAA gave the smallest mean attitude error. A claim of generally superior PFAA alignment is not supported.

\subsection{Irregular-wave terminal audit and distinct timing protocols}
The irregular-wave deck audit (W14.3) evaluated a block-input baseline, a fixed Haar basis and a deck-informed Haar basis under an illustrative irregular-wave realization. Table~\ref{tab:w14wave} reports errors relative to the moving clearance-offset reference. The very small terminal errors do not establish touchdown, and the three controllers should not be interpreted as a statistical sample of different sea states.

\begin{table}[t]
\centering\caption{W14.3 wave-deck terminal audit; not the Transformer benchmark.}
\label{tab:w14wave}
\resizebox{\columnwidth}{!}{%
\begin{tabular}{lrrr}\toprule
Metric & Block & Fixed Haar & Deck Haar\\\midrule
Position error (mm)&0.163&0.140&0.145\\
Relative speed (m/s)&0.002751&0.002714&0.002722\\
Attitude error (deg)&0.0389&0.0373&0.0378\\
Minimum clearance (m)&0.199880&0.199891&0.199891\\
Actuator violations&0&0&0\\\bottomrule
\end{tabular}}
\end{table}

The W13.5 single-update timing study used three repetitions and reported median block/fixed-Haar/deck-Haar times of $9.940/7.003/8.859$~s, but means of $8.313/8.527/8.059$~s. The W14.1 20-update comparison, performed under a different protocol, reported fixed/deck-informed mean update times of $4.102/2.422$~s with 20/20 nominal deadline misses in both cases; these are means, not the medians quoted below. Neither set can be pooled with the first-update Short10 or CL-33 acceleration trials.

\subsection{Original Transformer proposal diagnostics}
The autoregressive-proposal diagnostic (T03.7i) compared the original return-to-go (RTG) initialization with a horizon-aware alternative at updates $k\in\{0,1,5,10,15,19\}$. Mean thrust RMSE against expert controls was $1.4782$~N for the original conditioning and $1.6097$~N for horizon-aware conditioning. The tested modification therefore did not improve the measured proposal error.

In the separate T03.8 open-loop physical rollout of the original Transformer proposals, terminal position errors at those same checkpoints were $26.5614$, $18.9474$, $11.1658$, $0.3850$, $0.0068$, and $0.0056$~m. The corresponding attitude discrepancies against expert rollouts were $149.86^\circ$, $175.78^\circ$, $120.21^\circ$, $2.88^\circ$, $4.31^\circ$, and $2.07^\circ$. No predicted actuator component saturated in that audit. These are \emph{proposal-only} rollout diagnostics, not optimized closed-loop errors; they demonstrate that unsaturated controls may still yield poor terminal states. They do not uniquely identify the source of the error.

\subsection{Full- versus short-decision-horizon benchmark detail}
\begin{table}[t]
\centering\caption{Matched 30-scenario first-update comparison (median unless stated).}
\label{tab:ht10detail}
\resizebox{\columnwidth}{!}{%
\begin{tabular}{lrr}\toprule
Metric & Full20 Haar & Short10 v2\\\midrule
Feasible scenarios&29/30&13/30\\
Position error (m)&0.0115&0.0249\\
Position 90th percentile (m)&0.0759&0.1091\\
Relative speed error (m/s)&0.0124&0.3142\\
Wall time (s)&8.8355&6.1248\\
Wall-time 90th percentile (s)&18.8143&10.0420\\
Deadline misses&30/30&30/30\\\bottomrule
\end{tabular}}
\end{table}
The Short10 acceptance procedure tests the full physical trajectory objective before and after each proposed correction. A decreasing objective for accepted iterations is a \emph{within-run} property and does not imply that the short-decision-horizon controller has a smaller objective than the full-horizon controller. The saved 30-scenario summary provides terminal metrics and timing, not a paired distribution of the full physical objective across the two methods. The different iteration caps also prevent interpreting the observed time difference as a pure effect of horizon length.

\subsection{Original Haar and Transformer closed-loop benchmarks}
A separate 20-update Transformer-versus-shifted-initialization comparison must not be conflated with the later CL-31--33 acceleration experiments. In that separate experiment, Transformer initialization reduced mean recorded wall time from $3.9375$ to $1.4039$ s, but degraded final position error from $0.000801$ to $0.113755$ m and attitude error from approximately $0.275$ to $45.400$ degrees. Terminal relative-speed error increased from $0.002952$ to $0.372377$ m/s. The faster learned initialization therefore did not provide acceptable equivalent tracking in that comparison; neither implementation reliably met the nominal $0.2$-s deadline.

A separate W14 20-update fixed-versus-deck-informed Haar test recorded median wall times $3.538$ and $2.216$ s, with nearly identical terminal tracking in its sampled scenario. A distinct three-repetition W13.5 timing comparison recorded block, fixed-Haar and deck-informed-Haar medians of $9.940$, $7.003$ and $8.859$ s, respectively, while their means were $8.313$, $8.527$ and $8.059$ s. These differences demonstrate timing variability; they are not pooled into a single population-level speedup estimate. All recorded deadlines in the 20-update fixed/deck-informed experiment were missed. The choice of basis can therefore affect the computational cost without appreciably changing terminal tracking in this case. For practical implementation, however, the computation must also be completed within the prescribed update interval.\par\vspace{0pt plus 1fill}

\subsection{Adaptive-basis single-scenario pilot}
A separate initial-update experiment compared a fixed 13-column Haar basis against a trajectory-informed 13-column selection for one irregular-wave scenario (seed 300001, $H_s=0.8414795$ m and $T_p=10.8792402$ s). Both used eight SCP iterations. The fixed basis achieved $0.00203$ m terminal position error and $0.00159$ m/s relative-speed error in $29.357$ s of recorded SCP time; the adaptive basis achieved $0.06460$ m and $0.06121$ m/s in $17.542$ s. Both satisfied the illustrative terminal thresholds, but the adaptive method sacrificed accuracy for lower recorded optimization time. The timing excludes the basis-selection overhead. This pilot is not evidence of a general advantage across operating conditions.

\subsection{FAA/PFAA comparison across scenario sets}
The four-scenario FAA05 pilot yielded mean position errors of $0.03602$, $0.03571$ and $0.03469$ m for fixed, FAA and PFAA weights. Their mean attitude errors were $1.82271$, $1.81328$ and $1.82019$ degrees, and late attitude RMS errors were $1.33010$, $1.32561$ and $1.32975$ degrees. The subsequent 30-scenario FAA06 comparison, which is the stronger evidence base, found 13/30 predicted feasible cases for all three variants. Its mean late attitude RMS errors were $1.3849$, $1.3794$ and $1.3851$ degrees. Thus the modest position improvement under PFAA is not accompanied by an observed feasibility improvement, while FAA gives the smallest sampled attitude error. The pilot and 30-scenario results are reported separately.

\subsection{Numerical verification of acceleration methods}
Before the closed-loop comparisons, a condensed versus recursive convex-subproblem test found objective agreement on the order of $10^{-8}$ and maximum input and state discrepancies $3.51\times10^{-5}$ and $6.54\times10^{-6}$, respectively. In a three-iteration full-horizon comparison, both physical objectives were approximately $33.00838$, with an absolute difference of $5.71\times10^{-7}$, with wall times $6.025$ and $5.692$ s. A derivative-only test reported approximately a $2.04$-fold reduction in Jacobian assembly time, with maximum differences $1.08\times10^{-5}$ in $A$ and $8.81\times10^{-7}$ in $B$. A separate one-update forward-difference test recorded wall times $6.688$ and $3.511$ s. These numerical tests verify individual computational modifications. Their combined effect is assessed separately through matched closed-loop simulations.

\subsection{Independent matched closed-loop acceleration tests}
The CL31 derivative comparison, CL32 sensitivity-condensation comparison and CL33 combined-acceleration comparison use the same scenario, Transformer72 initialization, PFAA settings and 20-update shrinking-horizon protocol. They are \emph{single-scenario} closed-loop comparisons. Each nominal update advances the plant by $0.2$~s even when optimization wall time is much larger.

\begin{table*}[t]
\centering\caption{Measured matched 20-update closed-loop computation times. The final no-optimization update is excluded from time statistics. Reference timing varies across separate Colab runs.}
\begin{tabular}{lrrrr}\toprule
Experiment and comparison&Mean (s)&Median (s)&Max (s)&Deadline misses\\\midrule
CL-31 original central&4.158&4.495&8.485&19/19\\
CL-31 original forward&3.284&3.842&6.563&19/19\\
CL-32 original central&3.971&5.039&7.961&19/19\\
CL-32 condensed central&2.733&3.039&5.811&19/19\\
CL-33 original central&4.152&4.771&7.809&19/19\\
CL-33 condensed forward&2.030&2.153&3.913&18/19\\\bottomrule
\end{tabular}
\label{tab:times}
\end{table*}

The matched CL-33 median computation time decreased from 4.771 to 2.153 s, a 54.9\% reduction. The mean computation time decreased by 51.1\%. These figures measure elapsed wall-clock time for the complete timed controller update, not just the convex solver. The measurement includes the modeled trajectory evaluations and derivative calculations performed inside that update. Because the simulation advances by $T_s=0.2$~s irrespective of this measured duration, it is not a real-time hardware-in-the-loop validation. Separate-run speedups are not multiplied.

\begin{table*}[t]
\centering\caption{CL-33 original versus combined condensed/forward closed-loop terminal metrics.}
\begin{tabular}{lrr}\toprule
Metric&Original&Combined\\\midrule
Terminal position error (m)&0.0003493700605&0.0003493709949\\
Terminal velocity error (m/s)&0.002237597614&0.002237603039\\
Terminal attitude error (deg)&0.293976508&0.293976673\\
Terminal angular-rate error (rad/s)&0.00293402053&0.00293401981\\
Accepted SCP iterations&27&27\\
Rejected SCP attempts&80&80\\\bottomrule
\end{tabular}
\end{table*}
The maximum absolute state-component difference in CL-33 was $1.42684\times10^{-5}$ and the maximum absolute control-component difference $4.77287\times10^{-6}$, in their respective vector-component units. These values indicate close agreement in the tested run, not exact equivalence in all scenarios. CL-31 independently compared central and forward derivatives; CL-32 independently compared original and condensed SCP; CL-33 tested both accelerations together.

\subsection{Computational acceleration and real-time limitations}
Sensitivity condensation eliminates the explicit state-correction variables by applying the same linearized state recursion. This can reduce the size of the convex program but requires building the sensitivity matrices. Forward differences reduce the number of perturbed dynamics evaluations but may change numerical derivatives. The CL-31 and CL-32 tests isolate these modifications, while CL-33 measures their combined closed-loop effect. Because the original-reference timing differs across independent runs, the separate improvements cannot be multiplied to infer a combined gain. The CL-33 result instead provides a directly matched ratio under one scenario.

A particularly important negative observation is that the final optimization updates become cheaper as the horizon shrinks, yet 18/19 combined CL-33 updates still miss the prescribed $0.2$-s interval. The near-identical terminal errors demonstrate numerical consistency for that scenario under the same idealized propagation schedule; they do not establish stability under real computational latency. A future delay-aware experiment must advance the plant during computation or otherwise model the control hold and state-estimation age.

\section{Discussion and limitations}
The main numerical finding concerns computation: sensitivity condensation and forward differences reduce measured wall time while preserving the sampled closed-loop terminal tracking behavior in one matched scenario. The 30-scenario initial-update tests demonstrate that short-decision-horizon savings can come with substantially worse satisfaction of the relative-velocity requirement. The training-size experiment shows that lower action-prediction error does not necessarily improve satisfaction of the terminal tracking requirements.

The nominal $T_s=0.2$~s period is a design choice, not a universal operational landing requirement. Nevertheless, it is the specified update period for these simulations, and missed deadlines must be reported. In CL-33, 18 of 19 optimized updates exceeded this period; the plant simulation did not hold or delay controls by the measured solver time. The present study thus demonstrates neither delay-aware closed-loop performance nor onboard real-time operation. Increasing the controller period to accommodate multi-second optimization would change the four-second landing problem rather than retrospectively validate these results.

\subsection{Scope and experimental limitations}
The present investigation is a simulation-based study of the computational and terminal-tracking properties of Transformer-guided SCP for multirotor landing on a moving offshore deck. No physical flight experiments have been performed. The model assumes availability of the simulated vehicle state and prescribed deck-motion information; wind disturbances, measurement noise, estimator dynamics, motor response delays and touchdown contact are not represented. These assumptions allow controlled comparison of optimization formulations, but do not establish robustness to disturbances or imperfect measurements. The method is not claimed to inherit convergence, recursive feasibility or robustness guarantees from general SCP or MPC references.

\subsection{Experimental validation using measured data}
Future experimental work should replace idealized state and deck-motion information with synchronized UAV and landing-platform measurements. Candidate instruments include an onboard inertial measurement unit (IMU), relative-position sensing through vision or GNSS/RTK, and deck-mounted inertial and motion sensors. Measurements would enable assessment of deck-motion forecast error, state-estimation error, communication latency and their effects on terminal tracking. The present total-thrust and body-torque commands would require allocation to individual rotors, including motor dynamics, saturation and actuator response. A fiber-optic communication link or tethered experimental platform could be investigated, provided any mechanically significant tether forces and constraints are represented; a tethered vehicle is not dynamically identical to the present untethered model. Wind gusts and vessel-induced aerodynamic effects, as well as delay-aware computation and contact dynamics, should be examined before progressing from controlled laboratory tests to offshore flight experiments. The selection of a state estimator or uncertainty-aware control formulation should follow measured disturbance and sensing characteristics rather than being prescribed here. A low-tilt landing-window selector and offline trajectory library are separate future alternatives, not implemented in this study.

\subsection{Offline trajectory-library alternative and landing-window selection}
The observed deadline violations motivate a possible alternative architecture that was not implemented or benchmarked. One could solve many landing approaches offline and index the resulting trajectories by the relative state and deck forecast,
\begin{equation}
\xi=\bigl[r^\top,\ \dot r^\top,\ e_R^\top,\ e_\omega^\top,\ \widehat s^\top,\ T_{\mathrm{rem}}\bigr]^\top,
\end{equation}
where $\widehat s$ summarizes the predicted deck motion and $T_{\mathrm{rem}}$ is the remaining landing time. A library entry $\mathcal L(\xi_i)$ would store a nominal state trajectory, a thrust--torque sequence and feasibility diagnostics. A retrieval step could then supply $U_{\mathrm{lib}}$ to a short physics-based SCP correction:
\begin{equation}
U_{\mathrm{online}}=
\operatorname{SCP}_{N_{\mathrm{short}}}
\bigl(x_k,\widehat s_{k:k+H_s};U_{\mathrm{lib}}\bigr).
\end{equation}
This equation specifies a possible future architecture, not an implemented controller. Library coverage, interpolation error, changes in initial state, forecast uncertainty, safe-set conditions and a bounded-time fallback would require independent validation. In particular, a retrieved trajectory would not certify touchdown contact.

A separate future supervisory problem is to predict intervals of low roll, pitch, heave speed and deck angular rate, and select the terminal time $T_F$ before initiating tracking. This would change the fixed-terminal-time formulation studied here into a joint landing-time and control problem. Neither preferential low-tilt window selection nor its performance benefit is demonstrated in the present results.

\section{Reproducibility and scope of the results}
The reported first-update, attitude-weighting and acceleration studies use distinct experimental protocols and must not be pooled. The FAA/PFAA implementation uses the same nominal initial control proposal across its three modes, and its convex objective includes the attitude-stage penalty absent from the original baseline physical objective. The 54.9\% reduction in CL-33 median computation time is a matched single-scenario observation; no inference is made about a general distribution of sea states, hard real-time operation, or physical touchdown. These distinctions are necessary to reproduce the reported comparisons without combining incompatible timing protocols.

Representative experiments were rerun in a fresh Google Colab session with the archived project modules, the frozen Transformer checkpoint and the saved datasets. The selected short-decision-horizon test and the fixed/FAA/PFAA comparison completed successfully. For the matched combined-acceleration closed-loop scenario, the repeated maximum state-component and control-component differences were $1.4243\times10^{-5}$ and $4.7642\times10^{-6}$; the observed median computation-time reduction was approximately 37\%, compared with 54.9\% in the archived run. The difference in measured timing is expected to depend on runtime conditions and does not alter the original benchmark table. These representative checks do not replace the full 30-scenario datasets or establish onboard real-time execution.

\section{Conclusions}
This study examined Transformer-initialized, Haar-parameterized sequential convex programming within a model predictive control (MPC) framework for multirotor landing on a moving offshore deck. A key strength of MPC for this problem is its ability to optimize the approach over a finite prediction horizon while representing vehicle dynamics, a time-varying deck reference, actuator bounds and terminal tracking objectives in one constrained formulation. This is particularly useful when the landing target moves in heave, roll and pitch: the control decision can be informed by predicted motion rather than solely by the instantaneous tracking error. Combining a physics-based predictor with learned initialization and compact control coordinates also offers a systematic way to study computational cost against terminal accuracy. These are advantages of the formulation, not evidence that this controller outperforms PID, LQR or other architectures, which were not benchmarked here.

The principal findings are:
\begin{enumerate}
\item \textbf{Predictive landing formulation.} The constrained SCP--MPC formulation tracked a moving terminal reference with small position and relative-velocity errors in selected numerical scenarios. The simulated approach terminates at a prescribed clearance above the deck; neither physical touchdown nor flight performance was evaluated.

\item \textbf{Computational acceleration.} Sensitivity condensation and forward-difference derivatives reduced measured optimization-update wall time. In the matched CL-33, single-scenario, 20-update comparison, their combination achieved a \textbf{54.9\% reduction in median computation time} while producing nearly identical terminal errors and the same accepted SCP iteration count. This is the strongest directly matched computational improvement established here, but it is not a general estimate of computational savings.

\item \textbf{Transformer initialization and Haar coordinates.} Learned proposals and Haar control coordinates were integrated with physics-based SCP. The experiments characterize their behavior rather than proving an independent, universal advantage: increasing the Transformer training set did not improve the short-horizon feasibility count, and the Haar timing comparisons did not yield a uniform advantage across both mean and median times.

\item \textbf{Control-decision-horizon trade-off.} Shortening the adjustable decision horizon reduced computational work but lowered the fraction of cases meeting the illustrative terminal feasibility thresholds. Thus, computation time and terminal position and relative-velocity requirements must be assessed together.

\item \textbf{Attitude-weighting alternatives.} The fixed, FAA and PFAA variants produced modest differences in terminal errors, but each achieved \textbf{13/30 feasible cases} in the matched scenario set. No improvement in the feasibility count was demonstrated.

\item \textbf{Real-time and experimental limits.} Most timed updates exceeded the specified \textbf{0.2-s} control interval, including 18 of 19 combined CL-33 updates. The simulations also exclude physical flight tests, wind disturbances, measurement and estimation errors, motor-response delays and touchdown contact. Consequently, onboard real-time feasibility and operational offshore landing remain unvalidated.
\end{enumerate}

Overall, the results demonstrate a measured computational improvement for the tested SCP--MPC implementation while preserving closely matched simulated terminal tracking in the acceleration comparison. The contribution is a quantified assessment of the trade-offs among computational acceleration, decision-horizon length and terminal feasibility, not a claim of superior performance against other control families or experimentally validated landing. The reported experiments do not isolate every causal contribution of Transformer initialization, Haar-coordinate choice and horizon reduction in one factorial benchmark. Flight validation using measured data, actuator-level control allocation, disturbance assessment and delay-aware computation are appropriate next steps.

\end{document}